\documentclass[12pt,preprint]{elsarticle}
\usepackage[normalem]{ulem}
\usepackage{graphicx}
\usepackage{subcaption}
\usepackage{amsmath}
\usepackage{caption}
\usepackage[percent]{overpic}
\usepackage[table]{xcolor}
\usepackage{textcomp}
\usepackage{float}
\usepackage{times}
\usepackage{soul}
\usepackage{mathtools}
\usepackage{tabularx}
\usepackage{booktabs}
\usepackage{svg}
\usepackage{amssymb}
\usepackage{gensymb}
\usepackage[capitalize]{cleveref}
\usepackage{algpseudocode}
\usepackage{threeparttable}
\usepackage[final]{changes}

\begin{document}
\begin{frontmatter}

\title{Scaling Kinetic Monte-Carlo Simulations of Grain Growth with Combined Convolutional and Graph Neural Networks}

\author[1]{Zhihui Tian\fnref{fn1}}
\author[2]{Ethan Suwandi\fnref{fn1}}
\author[3]{Tomas Oppelstrup}
\author[3]{Vasily V. Bulatov}
\author[1]{Joel B. Harley}
\author[3]{Fei Zhou\corref{cor1}}

\address[1]{Department of Electrical and Computer Engineering, University of Florida, Gainesville, FL, USA}
\address[2]{Northwestern University}
\address[3]{Lawrence Livermore National Laboratory, Livermore, CA, United States}

\fntext[fn1]{These authors contributed equally to this work.}
\cortext[cor1]{Corresponding author.}
\ead{zhou6@llnl.gov}

\begin{abstract}
Graph neural networks (GNN) have emerged as a promising machine learning method for microstructure simulations such as grain growth.
However, accurate modeling of realistic grain boundary networks requires large simulation cells, which GNN has difficulty scaling up to. To alleviate the computational costs and memory footprint of GNN, we propose a hybrid architecture combining a convolutional neural network (CNN) based bijective autoencoder to compress the spatial dimensions, and a GNN that evolves the microstructure in the latent space of reduced spatial sizes. Our results demonstrate that the new design significantly reduces computational costs with using fewer message passing layer (from 12 down to 3) compared with GNN alone. The reduction in computational cost becomes more pronounced as the spatial size increases, indicating strong computational scalability. For the largest mesh evaluated ($160^3$), our method reduces memory usage and runtime in inference by 117× and 115×, respectively, compared with GNN-only baseline. More importantly, it shows higher accuracy and stronger spatiotemporal capability than the GNN-only baseline, especially in long-term testing. Such combination of scalability and accuracy is essential for simulating realistic material microstructures over extended time scales. The improvements can be attributed to the bijective autoencoder’s ability to compress information losslessly from spatial domain into a high dimensional feature space, thereby producing more expressive latent features for the GNN to learn from, while also contributing its own spatiotemporal modeling capability. Training data are generated from stochastic grain growth simulations, providing realistic variability for learning robust microstructure evolution. Comprehensive system validation confirms that the model is accurate, robust, and scalable.
\end{abstract} 


\end{frontmatter}

\section{Introduction}
\label{sec:introduction}

Polycrystalline materials comprise the bulk of modern engineering materials. Many physical properties of these materials, e.g.\ mechanical, chemical, electrical, magnetic, are intrinsically linked to their grain microstructure, i.e. size, shape and topology of crystal grains comprising the material \cite{Thompson2000ARMS, Suryanarayana1995IMR}. Therefore, understanding and predicting the evolution of grain microstructures during processing is important for developing and optimizing new materials with properties tailored for specific applications. 

The evolution of grains is driven by a complex variety of mechanisms at multiple length scales \cite{Thompson2000ARMS, Rios2018MST}. 
Several computational approaches have been used so far to simulate the coarsening of grain structures \cite{ Anderson1984AM, Srolovitz1984AM, Frost1996COSSMS}, such as molecular dynamics (MD)\cite{frenkel2023understanding},  kinetic Monte-Carlo (KMC) \cite{anderson1989kmc,  wright1997potts, zollner2006kmcstats, oppelstrup2016spock}, phase-field methods \cite{Fan1997AM, Kazaryan2001PRB,Kim2006PRE, Chen2002ARMR}, and finite-element front-tracking methods \cite{Frost1988SM, Fayad1999SM, lazar2011more}, roughly in order of increasing granularity and decreasing levels of detail, with each offering a unique balance between computational efficiency and physical fidelity. Each of these techniques is built on a set of physical assumptions and are concerned with capturing different aspects of microstructure evolution. At the finest level of detail, molecular dynamics is considered the most predictive. However, the computational resources required to simulate grain growth with MD are prohibitively high due to the length and time scales. On the opposite end of the spectrum, phase-field and front-tracking methods, coarse-grained models based on simplifying assumptions and continual representations, can reach much larger length and time scales. Although such continuum methods have contributed significantly to our understanding of grain growth mechanisms, the trade-off is often a reduced connection to physical details \cite{Chen2002ARMR}.

Potts model kinetic Monte Carlo (PMC) \cite{anderson1989kmc,  wright1997potts, zollner2006kmcstats, oppelstrup2016spock} is a stochastic microstructure simulation technique widely used due to its ease of implementation and scalability. In PMC, the grain structure is represented as a grid of lattice sites labeled with grain orientations, and each grain consists of a contiguous region or cluster of sites sharing the same grain label. The interfacial energy at grain boundaries (GB) is approximated with a Potts model that penalizes dissimilar labels at neighboring sites. 
In PMC simulations, a sequence of randomly selected lattice sites and label changes is proposed, each change either accepted or rejected in accordance with the Boltzmann distribution, leading to reduction of excessive interfacial energy associated with label/orientation mismatch between adjacent grains. 
As a result, the average grain size increases and the GB area per unit volume decreases, entailing macroscopic grain growth or coarsening.

PMC enjoys benefits from both ends of the spectrum: compared to MD, PMC is orders of magnitude more computationally efficient because of spatial coarse-graining of multiple atoms into a single lattice site and temporal coarse-graining of atomic vibrations into collective grain boundary motion in the form of label-flipping events. Compared to continuum methods, lattice sites in PMC retain discrete degrees of freedom with an energy model designed to emulate the effects of realistic atomic interactions. With properly parametrized Potts models, PMC has the capacity to simulate a wide and rich variety of grain coarsening mechanisms \cite{rodgers2015kmcAM} in agreement with established theories of grain growth \cite{von1952metal, mullins1956two, ullah2017kmc}. The kinetic Monte Carlo algorithm in SPOCK can efficiently scale up to $10^{11}$ elements (spins) in massively parallel 3D simulations in a heroic demonstration, although doing so required 1.5 million compute cores\cite{oppelstrup2016spock}. Maintaining such capabilitiesfor large and, simultaneously, detailed simulations while reducing their computational costs is therefore a highly attractive goal.

Scientific machine learning (ML) approaches have emerged as a fast-growing field of enabling techniques that complement and extend the capabilities of traditional computational materials science methods \cite{Butler2018N, Zhang2023-AI4Science}.
%{Schmidt2019,Wei2019,Butler2018,Damewood2023}
In particular, ML-based surrogate models trained on high-fidelity but expensive simulators such as PFM or PMC can accelerate microstructure evolution simulations with reduced computational costs \cite{Yang2021P,MontesdeOcaZapiain2021nCM, Wu2023CMS, KazemzadehFarizhandi2023CMS, Ji2025-Scalable}. 
A range of microstructure evolution phenomena, including spinodal decomposition \cite{Yang2021P,MontesdeOcaZapiain2021nCM}, grain growth \cite{Yang2021P} and dendrite growth \cite{Yang2021P, Ji2025-Scalable}, have been successfully reproduced with convolutional neural networks (CNN) %and recurrent neural networks (RNN) 
based surrogate models trained from phase-field data. Enhanced computational efficiency was achieved with coarse-grained spatiotemporal grids in CNN compared to fine grids in the differential equation solvers of PFM \cite{Yang2021P,Ji2025-Scalable}. Although several studies have applied deep learning methods to the grain growth problem, most of them is based on phase field or other deterministic data\cite{Yang2021P,Fan2024MLST,peivaste2025teaching,oommen2024rethinking,qin2024graingnn}. Only one study uses a deep neural network to learn how the energy decreases with the physically informed regularization\cite{Yan2022MD}. Rather than inputting the microstructure directly into the network, their method firstly encodes it with the Hamilton function inspired by KMC, aiming to capture the energy decay dynamic. However, this approach employs a fixed window size for extracting the local information across time, which overlooks the dynamic nature of grain growth. For instance, the average grain area increases over time, which requires a correspondingly adaptive window and could be solved by a graph neural network with adaptive remeshing.

In the rapidly evolving field of scientific machine learning, 
graph neural networks (GNNs) \cite{Fan2024MLST,nino2025data,qin2024graingnn} have garnered considerable attention as a flexible ML architecture in problems with data structure that resembles computational graphs, such as particle-based fluid simulations \cite{Sanchez-Gonzalez2020, Pfaff2020GNN} and discrete dislocation dynamics \cite{Bertin2023JCP-Accelerating, Bertin2024nCM-Learning}. GNNs have proved to be more accurate than CNN in some grid/mesh based simulations of fluid dynamics \cite{Pfaff2020GNN}. Very recently\cite{Fan2024MLST}, some of us showed that GNN-based surrogate models for grain growth significantly outperform our previous CNN models \cite{Yang2021P} and were almost indistinguishable from the ground-truth 2D and 3D phase-field simulations. However, GNN is computationally expensive  and restricted to relatively small simulation cells\cite{Fan2024MLST}. %because long-range dependencies in the full microstructure graph require deep message-passing layers. 
This is at odds with realistic grain boundary structures, which are inherently complex, featuring irregular shapes, triple junctions, quadruple nodes and a broad distribution of grain sizes and orientations. Small simulation domains often fail to reproduce this heterogeneity, leading to biased or incomplete representations of coarsening dynamics. ML models therefore must be sufficiently scalable to capture the full spectrum of grain boundary topology and evolution. \added{However, scaling GNNs to larger simulation domains remains difficult because larger graphs impose substantial memory costs\cite{duan2022comprehensive}, and modeling long-range dependencies often requires many message-passing layers, increasing the risk of oversmoothing\cite{rusch2023survey}.}
%Encoding the whole microstructure into a latent space, then using deep learning algorithms to learn the latent dynamics could be an approach to solve that. 

In this work, we develop a hybrid architecture that 
compresses the spatial configurations into latent features with a CNN-based reversible autoencoder and performs temporal predictions in the latent space. \replaced{Our hybrid architecture naturally sidesteps both pathologies: the CNN encoder handles multi-scale spatial compression efficiently, while the GNN operates in a compact latent space where only 3 message passing layers suffice to capture all relevant interactions. }{It combines the capacity of CNN for efficient compression and feature learning, and GNN's ability for dynamical predictions.}
%CRevNet~\cite{yu2020autoencoder} to encode the microstructure into latent variables, followed by a backbone GNN~\cite{Fan2024MLST} that models the dynamics in latent space.
The main contributions of this study are as follows: 
(1) a bijective autoencoder for efficient latent feature learning; 
(2) a hybrid architecture that achieves high accuracy with reduced memory cost and runtime; 
(3) systematic validation against a classical stochastic grain growth model, Potts Monte Carlo with isotropic GB energy.

\section{Methods}
\label{sec:methods}

\label{sec:kmc}

\subsection{PMC Training Dataset}  
Grain coarsening trajectories with isotropic grain boundary energy were generated with the Spock code \cite{oppelstrup2016spock} and subsequently postprocessed to reduce stochastic boundary fluctuations while retaining large-scale grain boundary motion. We define a postprocessing operator $\mathcal{P}$ mapping the discrete KMC configuration  
$
s \in \mathbb{Z}^{N_1 \times \cdots \times N_d}
$  
to a continuous order-parameter field  
$$
\phi(r) = \mathcal{P}(s) \in \mathbb{R}^{n_1 \times \cdots \times n_d}, \quad n_k = N_k/S,
$$  
where $S$ is the spatial down-sampling ratio. The operator $\mathcal{P}$ consists of: (i) boundary extraction, (ii) block averaging with Gaussian smoothing, (iii) temporal averaging, and (iv) normalization. The result is a smooth field $\phi$ that approaches 0 at grain boundaries and 1 within grains, emphasizing collective interface motion.  

In practice, the spatial grid was reduced from $256^2$ to $64^2$ in 2D and from \replaced{$64^3$}{$128^3$} to $32^3$ in 3D ($S^d$ down-sampling). The final dataset comprises time series of coarse-grained configurations with shape  
$
(N_b, N_t, n_1, \dots, n_d, c),
$  
where $N_b=351$ simulations, $N_t=25$ time steps, $c=1$ order parameter channel, and $n_k=64$ (2D) or 32 (3D). Grid sizes were chosen to balance microstructural variability with sufficient resolution to represent grain boundaries by at least 1--2 pixels.

A Monte Carlo method, PMC is intrinsically stochastic. Fig.~\ref{fig:shared_grains} shows the divergence of KMC simulations in $32^3$ and $96^3$, which demonstrates the stochastic nature of grain growth. In practice, microstructural evolution is not purely deterministic but subject to thermal fluctuations, local irregularities, and probabilistic grain boundary movements \cite{humphreys2012recrystallization}. MC trajectories therefore contain not only the mean trend of boundary migration but also high-frequency variations, fluctuations in grain size distribution, and sample-to-sample variability. 

Training a surrogate on such data enables the model to capture both the average dynamics and the statistical variance of the process. This capability is particularly valuable when comparing against experiments, which inherently exhibit variability. By learning from stochastic datasets, the surrogate can go beyond deterministic curve-fitting and provide uncertainty-aware predictions that more faithfully reflect the physical system. Correspondingly, beyond the pixelwise or voxelwise metrics like RMSE for comparing the surrogate model and the ground truth data, it is more important that ML captures the statistical features of grain coarsening.

\subsection{Hybrid neural network architecture} 
\label{sec:NN}
{Realistic grain boundary networks require simulations with millions of lattice sites to capture long-range interactions and statistical variability. Training and inference on such large microstructures pose significant computational challenges, as the model must efficiently process complex, high-dimensional data while maintaining physical fidelity. Scalable learning architectures are therefore essential to handle the size and diversity of realistic grain growth datasets.}

\begin{figure}[htbp]
  \centering
  \begin{subfigure}{\linewidth}
    \includegraphics[width=\linewidth]{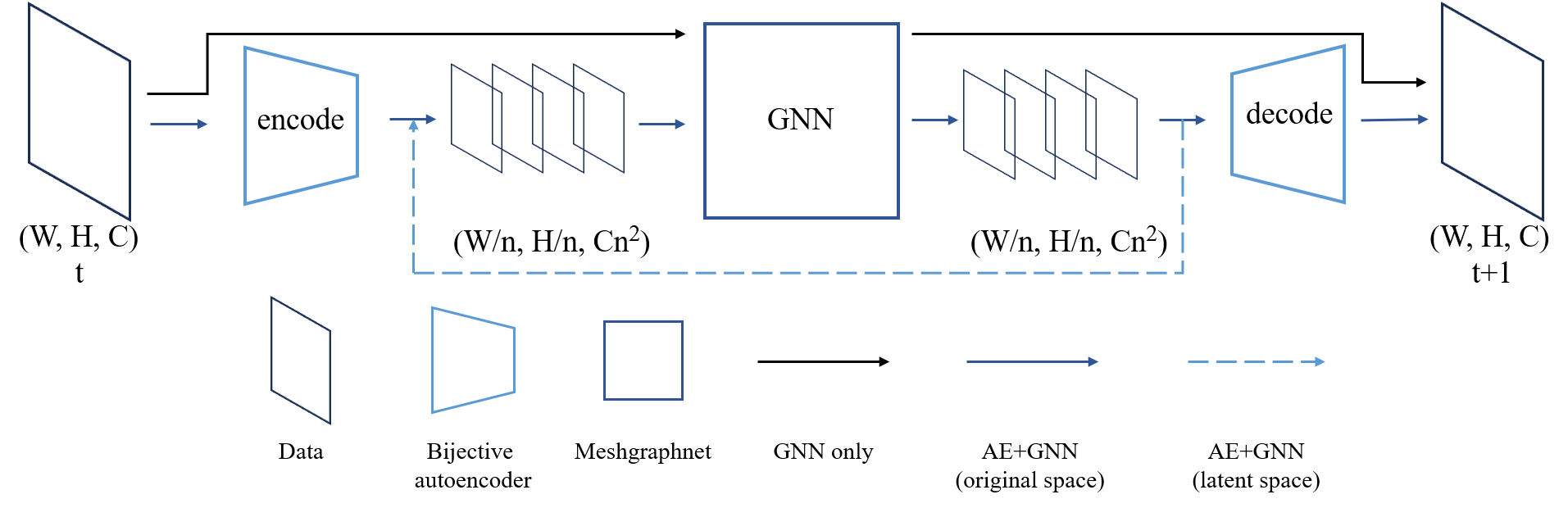}
    \caption{}
    \label{fig:NN-arch-a}
  \end{subfigure}

  \begin{subfigure}{\linewidth}
    \includegraphics[width=\linewidth]{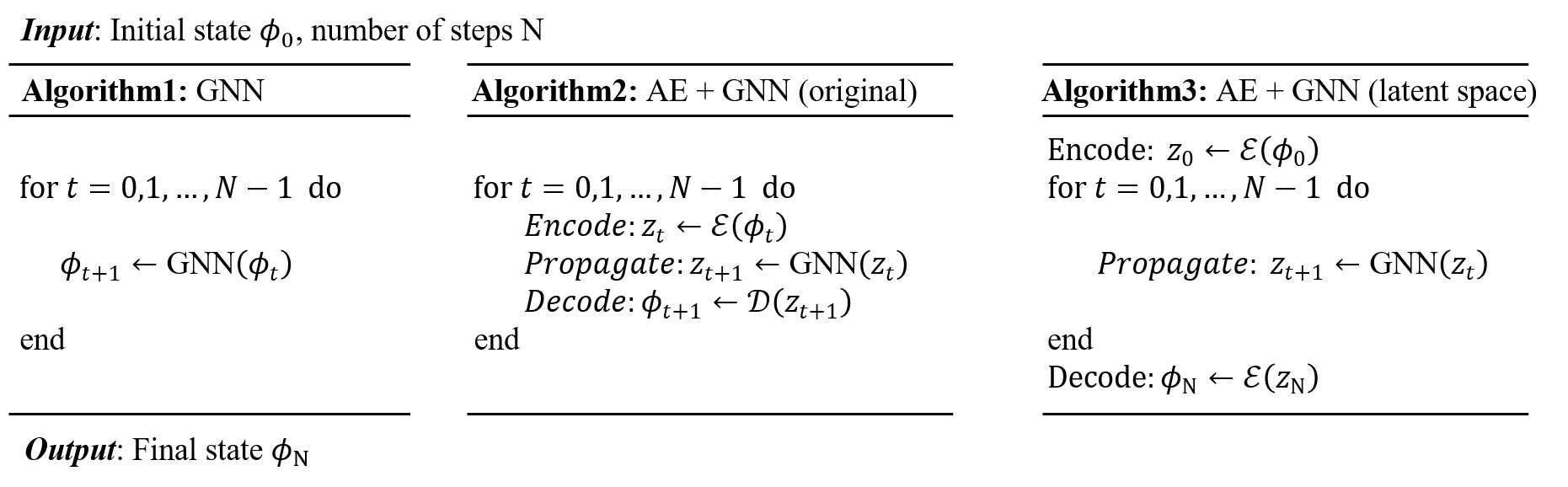}
    \caption{}
    \label{fig:NN-arch-b}
  \end{subfigure}

  \caption{
  Architecture of the ML models. 
  (a) shows the model structures, and (b) presents the corresponding pseudocode.
  The detailed network structure is given in Fig.~\ref{fig:NN-concept}.
  }
  \label{fig:NN-arch}
\end{figure}

% \textbf{Hybrid Architecture:}
As outlined in Figure \ref{fig:NN-arch}, we propose a novel hybrid architecture to simultaneously reduce memory footprint and computational cost: a CNN-based bijective autoencoder (AE) \cite{yu2020autoencoder} to reduce the dimension of the simulation cell with no information loss and a graph network \cite{Fan2024MLST} that performs time-evolution prediction in latent space.  All ML models were implemented in the PyTorch package at single (32 bit) floating-point precision with 96 hidden features. \replaced{
Due to the spatial compression introduced by the CNN encoder, the proposed AE+GNN architecture requires substantially fewer message passing layers than a pure GNN operating on the original grid. In this study, we evaluate pure-GNN baselines with 3, 5, and 12 message passing layers, with the 12 layer model providing the best performance among the tested configurations (Fig.~S3). Even compared with the strongest pure-GNN baseline, the proposed AE+GNN model achieves higher accuracy using only 3 message passing layers, enabling improved scalability without sacrificing predictive performance.
}{To improve scalability, we substantially simplify the GNN model from 10 layers in Ref.~\cite{Fan2024MLST} to 3 in this work.}

\textbf{Bijective autoencoder:} 
The AE is a frequently used tool for dimension reduction. It typically transforms an input $c$-dimensional variable $\phi \in \mathcal{R}^{c}$ into a bottle-neck latent feature vector $z=\mathcal{E}(\phi) \in \mathcal{R}^{c'}$ in a lower dimension $c'<c$, and then approximately recover the original $\phi \approx \mathcal{D}({z})$, where $\mathcal{E}$ ($\mathcal{D}$) are the encoder (decoder) part of the AE. Instead of commonly used lossy AE, we employ a two-way reversible autoencoder \cite{yu2020autoencoder} with a unified encoder-decoder design, which performs lossless spatial compression with $\mathcal{D}=\mathcal{E}^{-1}$. Pixel shuffle layer\cite{shi2016real}, a bijective downsampling is used to guarantee the invertibility of the entire AE. The encoder changes the shape of features from $(W, H, C)$ to $(W/n, H/n, C n^2)$ in 2D shown in Figure \ref{fig:NN-arch}(a) with the linear compression ratio $n$
%=1, 2, 4, 8$
and feature channel dimension expansion ratio $n^2$, such that the total data size
%product of the spatial and channel dimensions 
is preserved. In 3D, the feature is similarly tranformed from $(D, W, H, C)$ to $(D/n, W/n, H/n, Cn^3)$.

\textbf{Graph Neural Network:} 
We adopted the MeshGraphNet (MGN) model \cite{Pfaff2020GNN}, which has proved to be well suited as a surrogate model for graph-structured mesoscale materials simulations, including phase-field based grain growth \cite{Fan2024MLST} and dislocation dynamics \cite{Bertin2023JCP-Accelerating, Bertin2024nCM-Learning}. The graph based architecture allows for more flexible representations compared to CNN \cite{Yang2021P}. More implementation details can be found in our previous work \cite{Fan2024MLST}.

\textbf{Inference:} 
The guaranteed invertibility enables two inference strategies. The first works directly in the original data space ({solid blue line in} Figure \ref{fig:NN-arch}(a)):
\begin{equation}
\phi_{t+n} = F^{(n)}(\phi_t) 
\label{eq:ar1}
% \end{aligned}
\end{equation}
where 
$
F = \mathcal{D} \circ G \circ \mathcal{E}: \phi \to \phi
$ is the autoregressive predictor in the original data space and $G$ is a GNN in the latent space. The second algorithm (dashed line in Figure \ref{fig:NN-arch}a) takes advantage of the fact that the encoder and decoder are exact inverses of each other to simplify Eq.~(\ref{eq:ar1})
\begin{equation}
\phi_{t+n} =\underbrace{
\left(\mathcal{E}^{-1} \circ G \circ \mathcal{E} \right) \circ \cdots \circ \left( \mathcal{E}^{-1} \circ G \circ \mathcal{E} \right)
}_{n \text{ times}} (\phi_t)  = \mathcal{D} \circ G^{(n)} \circ \mathcal{E} (\phi_t).
\label{eq:ar2}
% \end{aligned}
\end{equation}

This eliminates the need for repeated decoding–encoding at each step,
thereby further reduces memory cost and, in particular, runtime, since the encoder and decoder are applied only once rather than at every step of the multi-step rollout, as illustrated by the comparison between Eq.~\eqref{eq:ar1} and Eq.~\eqref{eq:ar2}.
% The architecture from our previous work using only a GNN is also shown in Fig \ref{fig:NN-arch}, where the GNN directly evolves the system in the original space and can be represented as:
The GNN-only architecture without compression is also shown in Fig.~\ref{fig:NN-arch}a (top black arrow). The pseudocodes for the three algorithms are shown in Fig.~\ref{fig:NN-arch}b. {A detailed plot of the architecture can be found in the Supporting Information (Fig. S1).}

The computational advantage of the hybrid CNN--GNN framework can be understood from the spatial compression introduced by the CNN encoder. Specifically, the encoder reduces the spatial resolution by a factor $s$ in each spatial dimension, such that each node in the latent graph represents approximately $s^D$ grid cells of the original microstructure, where $D$ denotes the spatial dimension ($D=2$ for 2D and $D=3$ for 3D). Consequently, the effective node spacing increases from $\delta x$ to $s\,\delta x$. In a message passing neural network, the physical interaction range achieved after $L$ layers scales as $L \cdot \delta x$. After spatial compression, this becomes $L \cdot s \cdot \delta x$. To capture the same physical interaction range $R$, the required number of message passing layers is therefore reduced as
\begin{equation}
L_{\text{latent}} = \frac{R}{s\,\delta x} \approx \frac{L_{\text{original}}}{s}.
\end{equation}

In addition, the memory consumption $M$ of graph neural networks scales approximately with the number of nodes, $N_{\text{nodes}} \sim n^D$. After spatial compression, the graph size becomes $(n/s)^D$, leading to
\begin{equation}
M_{\text{latent}} \sim \frac{M_{\text{original}}}{s^D}.
\end{equation}

Overall, the hybrid architecture lowers the computational costs from three perspectives: 1. While the representations in the original space and latent space contain the same total number of features, spatial downsampling reduces the numbers of nodes and edges, which substantially decreases the memory and computational cost in the GNN \cite{Pfaff2020GNN}. 2. Moreover, in the reduced-dimensional representation, the receptive field can be captured with fewer message-passing layers, further improving efficiency. 3. By leveraging the reversibility of the autoencoder, performing the rollout directly in the latent space further reduces the computational cost without compromising accuracy.

Note that if a lossy compression is used instead, then the prediction accuracy and rollout stability will be affected, and latent space inference would not be feasible.

\subsection{Training of Neural Network}
\label{sec:training}
%Adhering to accepted NN training practice, 
The dataset was randomly partitioned into two subsets: training and validation. All models were trained on a single NVIDIA Volta V100 GPU with 16 GB of memory for smaller spatial domains or AMD MI300A APU with 100 GB of memory for larger spatial domains. %\replaced[id=ZT]{
The Adamw optimizer \cite{loshchilov2017decoupled} was adopted with the plateau learning rate scheduling method initialized at $10^{-3}$.  To improve both accuracy and generalizability, we designed specialized training procedures for the surrogate model:

(1) \textbf{Noise injection.}  
Following Refs.~\cite{Sanchez-Gonzalez2020, Pfaff2020GNN, Godwin2021}, we applied Gaussian noise to stabilize long-term rollouts. Surrogate models often accumulate errors when applied autoregressively, leading to instability. To mitigate this, we added small Gaussian perturbations $\epsilon \sim \mathcal{N}(0,1)$ with amplitude $10^{-3}$ to each training input frame, i.e.,
\[
\phi_t \;\mapsto\; \phi_t + 10^{-3}\epsilon.
\]
Noise was not added to validation or test data. This procedure teaches the model to tolerate small imperfections in its own predictions, resulting in more stable long-term dynamics. Similar denoising strategies have recently been successful in crystal structure classification across diverse solid phases~\cite{Hsu2024nCM-denoiser, Sun2024JCIM-ice}.

(2) \textbf{Multi-step self-supervised loss.}  
We evaluated different loss functions and adopted a mean-square ($L_2$) pixel-wise loss. To encourage the model to capture long-term dynamics, we employed a multi-step training scheme. The next-step prediction loss
$
L\left(\mathcal{F}(\phi_t), \phi_{t+1}\right)
$
was generalized to multiple autoregressive steps:
\begin{equation} \label{eq:multi-step-loss}
    L_\text{multi} =\sum_{k=1}^{p} L\!\left(\mathcal{F}^{(k)}(\phi_t), \phi_{t+k}\right),
\end{equation}
where p is a hyperparameter that influences the model long-term performance, and $\mathcal{F}^{(k)}$ denotes $k$ successive applications of the surrogate model. This reduces drift and improves stability over longer horizons.

(3) \textbf{Symmetry-based data augmentation.}  
To enforce invariance under rotational symmetry, we augmented training data using point-group operations. For 2D datasets, we applied all operations of the $4m$ group; for 3D, we used the cubic $O_h$ group. This augmentation forces the model to learn symmetry-equivariant dynamics \cite{Fan2024MLST}.

\section{Results and Discussion}
\label{sec:results}

\subsection{Performance} 
\label{sec:performance}

\begin{figure}[htbp]
  \centering
\includegraphics[width=\linewidth]{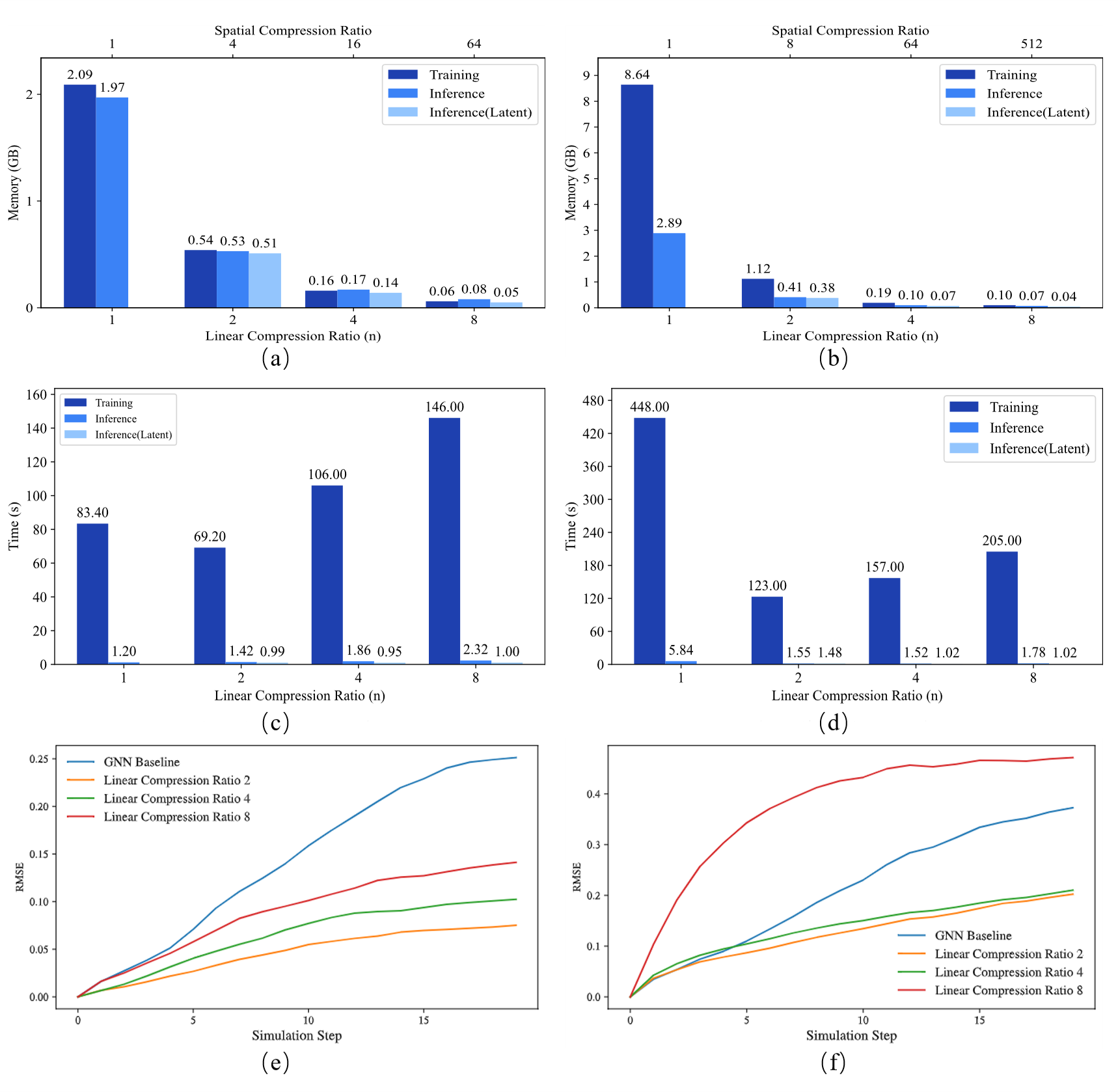}
  \caption{
 Effects of autoencoder (AE) compression ratios on training and inference computational costs and prediction accuracy {on an AMD MI300A APU with 3 message-passing layers in the GNN}. Memory usage comparison for (a) 2D system of $64^2$ mesh and (b) 3D with $32^3$ mesh; Run time for (c) 2D and (d) 3D; validation RMSE for (e) 2D and (f) 3D.}

\label{fig:memory}
\end{figure}

Figure \ref{fig:memory} shows the memory usage and runtime comparison between GNN-only and AE+GNN models, both using three message passing layers. In Fig.\ \ref{fig:memory}(a)–(d)
, a compression ratio of 1 corresponds to the original GNN model, which serves as the baseline for comparison. The linear compression ratios $n$ indicate the degree of spatial downsampling by the autoencoder applied for each dimension before passing the representation to the backbone GNN. Specifically, in two dimensions, linear compression ratios $=1, 2, 4, 8$ correspond to total spatial compression ratios of $n^2= 1, 4, 16, 64$, respectively. In 3D, they correspond to total compression total of $n^3= 1, 8, 64,  512$. The results show that our AE+GNN model  substantially reduces memory usage for both training and inference. Fig.~\ref{fig:memory}a,b shows reduced memory usage with increasing linear compression ratio $n$ for 2D and 3D. With $n=8$, training memory usage is reduced by approximately $35\times$ for 2D cases and $86\times$ for 3D cases. For runtime comparison in Fig.~\ref{fig:memory}c,d, we focus on the inference time. The AE+GNN model does not necessarily guarantee a shorter runtime than the GNN baseline, as the autoencoder introduces additional computational overhead. However, when inference is performed in the latent space (algorithm 3 in Fig.~\ref{fig:NN-arch}), the AE+GNN model achieves consistently lower runtimes compared to GNN, particularly at higher compression ratios. For example, with $n=8$ in 3D, the inference is approximately $6\times$ faster than using GNN only. 
The computational and runtime advantages of the bijective autoencoder become more pronounced in larger meshes. We run large mesh simulations on with AMD Instinct MI300A for 2D with and 3D with different large spatial sizes. Due to the GPU memory limitation of the GNN-only model, the largest simulation sizes used considered are $1280^2$ (2D) and $128^3$ (3D) for training, and $2688^2$ (2D) and $160^3$ (3D) for inference. For the largest training cases, in 2D, the proposed AE+GNN model reduces memory usage by approximately 54× and runtime by 12×. In 3D, the GNN-only model fails to run due to out-of-memory (OOM), whereas the AE+GNN model requires only a small amount of GPU memory. For the largest size cases during inference, memory usage is reduced by approximately 45× for 2D cases and 117× for 3D cases, accompanied by runtime reductions of about 43× and 115× respectively. {The large mesh inference memory and runtime comparison between GNN only, AE+GNN(original) and AE+GNN(latent) is shown in Tables~\ref{tab:3d_inference}. Detailed training comparisons across different spatial sizes are provided in Tables~\ref{tab:2d_train}–\ref{tab:3d_train} for 2D and 3D training, and in Table~\ref{tab:2d_inference} for 2D inference}. \added{A runtime comparison between the conventional KMC simulations and the machine learning based models is provided in Table S4. The proposed AE+GNN model significantly reduces the computational cost of microstructure evolution. For example, for the largest 3D case ($160^3$), the runtime is reduced from 809.9 s for the KMC simulation to 1.10 s using the AE+GNN (latent) model.}

\begin{table}[t]
\centering
\small
\setlength{\tabcolsep}{5pt}
\renewcommand{\arraystretch}{1.2}

\makebox[\textwidth][c]{%
\begin{threeparttable}
\begin{tabular}{l c c c c c c c}
\toprule
\textbf{Memory} & 
\textbf{GNN only} & \multicolumn{3}{c}{\textbf{AE+GNN(original)}}  & \multicolumn{3}{c} {\textbf{AE+GNN(latent)}} \\ 
\midrule
$96^3$ & 19.569
 & 2.802
 & 0.689
 & 0.484
& 2.528
& 0.414
&0.178
 \\

$128^3$ & 46.382
 & 6.616
 & 1.622
 & 1.133
& 5.970
& 0.976
&0.410
 \\
 
$160^3$ & \cellcolor{red!30}90.592
 & 12.929
 & 3.168
 & 2.208
& 11.666
& 1.904

&\cellcolor{blue!30}0.777
 \\
\midrule
\textbf{Runtime} & 
\textbf{GNN only} & \multicolumn{3}{c}{\textbf{AE+GNN(original)}}  & \multicolumn{3}{c} {\textbf{AE+GNN(latent)}}\\

\midrule
$96^3$ & 6.554e+00
 & 7.991e-01
 & 6.097e-01
 & 4.687e-01
& 8.380e-01
& 1.913e-01
&1.546e-01
 \\

$128^3$ & 1.270e+01
 & 1.919e+00
 & 1.314e+00
 & 8.116e-01
& 2.465e+00
& 8.036e-01
&7.581e-01
 \\
 
$160^3$ & \cellcolor{red!30}1.269e+02
 & 7.925e+00
 & 3.047e+00
 & 2.236e+00
& 7.057e+00
& 1.145e+00
& \cellcolor{blue!30} 1.104e+00
 \\

\bottomrule
\end{tabular}
\end{threeparttable}
}% end of makebox
\caption{Memory comparison(GB) and runtime(s) and  for large meshes in 3D inference. Columns under AE+GNN(original) and AE+GNN(latent) represent linear compression ratio 2, 4, 8 from left to right. Highlighted cells mark the largest mesh case, comparing the GNN-only baseline against the AE+GNN(latent) model with a compression ratio of 8.}
\label{tab:3d_inference}
\end{table}

Fig.~\ref{fig:memory}e, f compares the root-mean-square error (RMSE) of the baseline GNN and the AE+GNN models, both with 3 message-passing layers. The RMSE serves  as a simple pixelwise accuracy metric. More elaborate metrics  will be discussed later. It should be noted that the RMSE here is always higher than in our previous work \cite{Fan2024MLST}, because (i) the GNN model depth is reduced from 10 to 3 layers, and (ii) the current PMC data contain stochastic fluctuations that increase prediction errors compared to deterministic phase field data in Ref.~\cite{Fan2024MLST}. 
%, aiming to verify that learning ability is preserved while improving computational efficiency. 
In 2D (Fig.~\ref{fig:memory}e), the RMSE of AE+GNN is always lower than that of the baseline for all $n$ from 2 to 8. In 3D (Fig.~\ref{fig:memory}f), the cases with $n=2,4$ are more accurate than GNN but the $n=8$ model is worse. We attribute the improved accuracy at moderate $n=2,4$ to the increased receptive field of the GNN in the latent space, as each message-passing step can now transmit information from longer distance. The decreased accuracy at high compression ratio may be due to overfitting. Overall, the AE+GNN approach not only reduces memory consumption and runtime but also enhances predictive accuracy\added{, which may be partly attributed to the reduced spatial resolution of the encoded representation, since it allows long-range interactions to be captured with a shallower message-passing architecture and thereby lowers the risk of oversmoothing\cite{chen2020measuring, hu2019hierarchical}}.

\begin{figure}[htbp]
  \centering
\includegraphics[width=\linewidth]{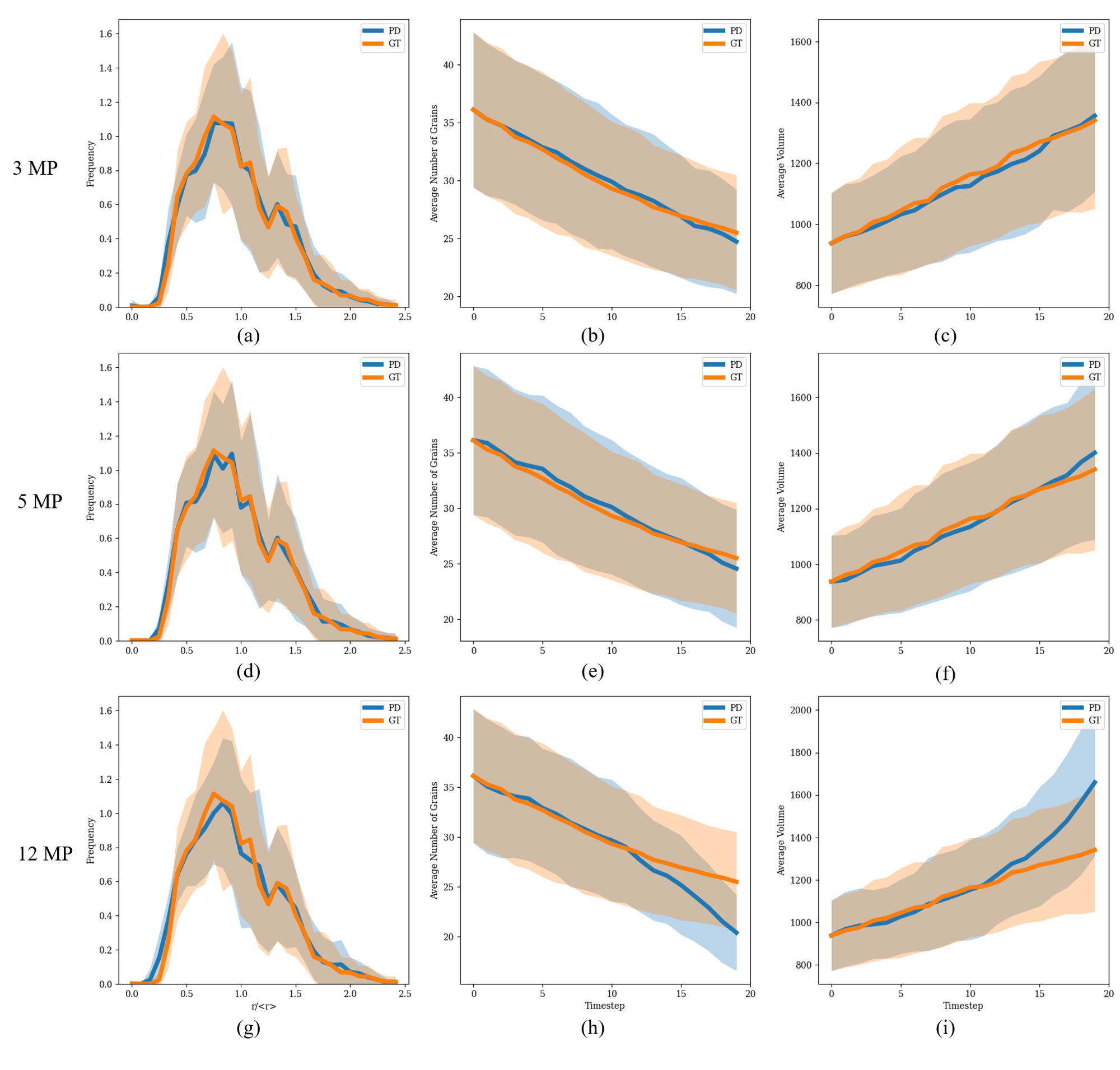}
  \caption{{Effects of number of message passing layer on AE+GNN model.} Statistical metrics of 3D grain simulations for AE+GNN based on 40 independent predicted and ground-truth trajectories using a $32^3$ mesh. The first, second, and third rows correspond to models using 3, 5, 12 message passing layers in GNN, respectively. From left to right, the columns show the normalized grain diameter distribution, the number of grains, and the average grain area as a function of time. The grain diameter distributions are computed by pooling grain statistics from timesteps $t=0$–$19$ across all samples, with grain diameters normalized by the mean diameter of each sample. Shaded regions denote one standard deviation across independent samples.
}
\label{fig:nmp}
\end{figure}

 Furthermore, this new architecture more effectively captures long-term dynamics using far fewer message-passing layers compared to the GNN-only model. In our previous study \cite{Fan2024MLST}, the MGN used 10 message-passing layers, which produced acceptable results but at the cost of computational efficiency. Excessive message passing also introduces the oversmoothing issue in GNNs.\cite{keriven2022not}. The effect of the number of message-passing layers, treated as a hyperparameter, is investigated to evaluate its impact on model performance. The results in Fig.~\ref{fig:nmp}(a-c) indicate that three layers are sufficient for this new architecture to ensure stable and accurate predictions, while further increasing the number of message passing layer leads to performance degradation shown in Fig.~\ref{fig:nmp}(g-i). For comparison, this hyperparameter is also examined in the GNN-only model (Fig.~\ref{fig:gnn_nmp}). A small number of message-passing layers fails to capture the PMC dynamics, whereas adding more layers yields moderately improved results at the expense of higher computational cost but still fails to reproduce the statistical characteristics of the ground truth. In contrast, the AE+GNN model requires only three message-passing layers to ensure both accuracy and efficiency. {The ability of AE+GNN to capture long-range dynamics more efficiently with fewer MP layer is attributed to the compressed latent space: information passed from 1 latent voxel away is actually propagated from $n$ real-space voxels away.}

\subsection{Extrapolation and visualization}

\begin{figure}[htbp]
  \centering
\includegraphics[width=\linewidth]{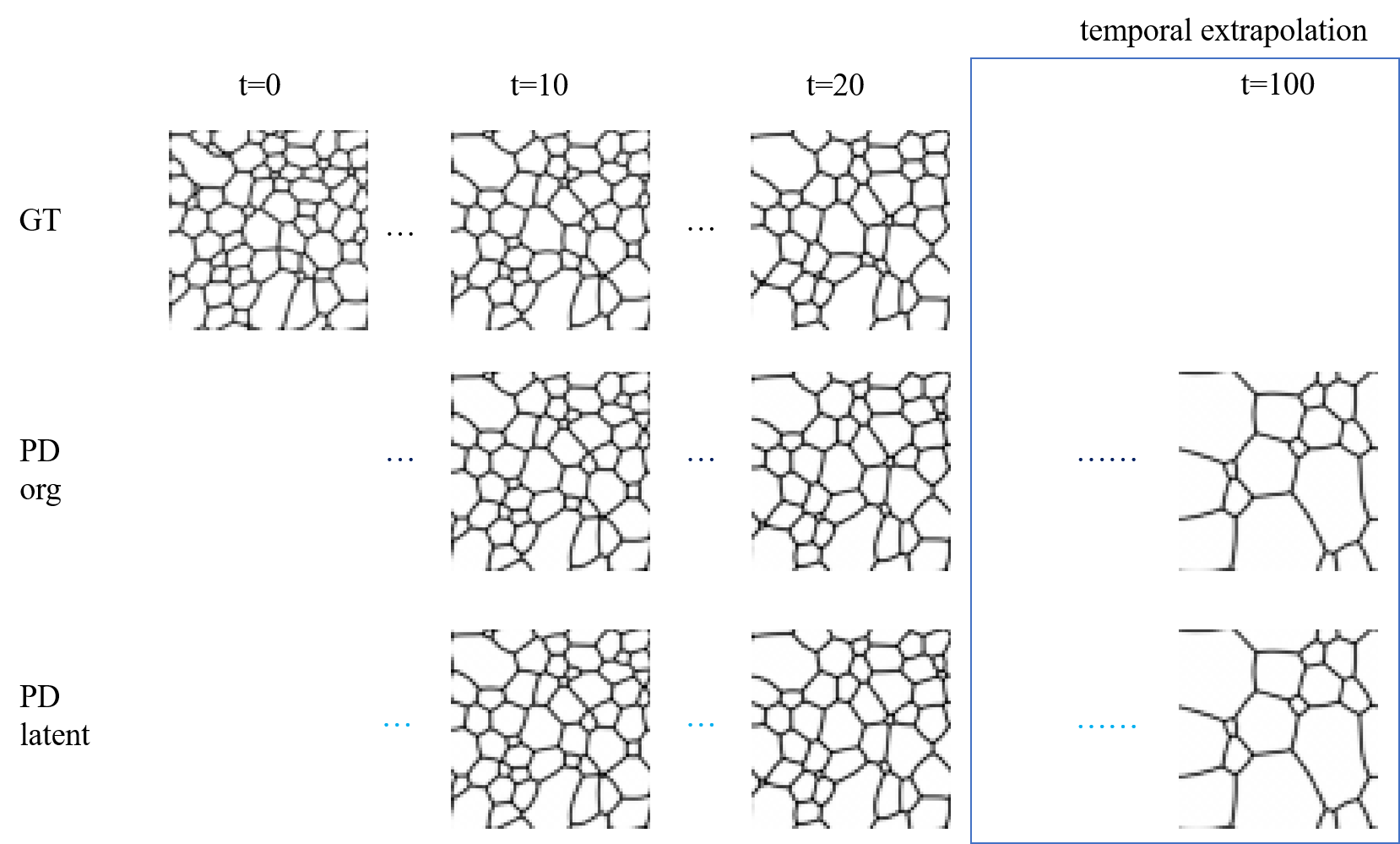}
  \caption{Temporal extrapolation and microstructure visualization of 2D predictions on a $64^2$ mesh (compression ratio $n=4$), trained trajectories with 25 frames and inferred for 100 frames in original and latent spaces {using 3 MP layer}, demonstrating temporal extrapolation.  The rows are, from the top, ground truth PMC data, predictions using algorithm 2, and algorithm 3 (latent-space inference). Ground-truth data are available only up to $t=25$; therefore, the frame at $t=100$ corresponds to an extrapolated prediction beyond the available ground truth.}
\label{fig:2dvisual}
\end{figure}

For 2D, the ground truth dataset consists of coarsened PMC trajectories each with $N_t=25$ frames of $64^2$ pixels, split into 351 trajectories for training and 123 for validation. The ground-truth PMC data were down sampled spatially by a factor of $4^2$ from the MC lattice of \replaced{$256^2$}{$1024^2$} and temporally by 8 from 200 frames. An exemplar trajectory 2D is shown in Figure~\ref{fig:2dvisual}. Training on 25 steps and inferring over 100 steps demonstrates the model’s temporal extrapolation capability. Furthermore, the latent-space prediction (algorithm 3) at different time steps is identical to the corresponding prediction of algorithm 2, as expected.

\begin{figure}[htbp]
  \centering
\includegraphics[width=\linewidth]{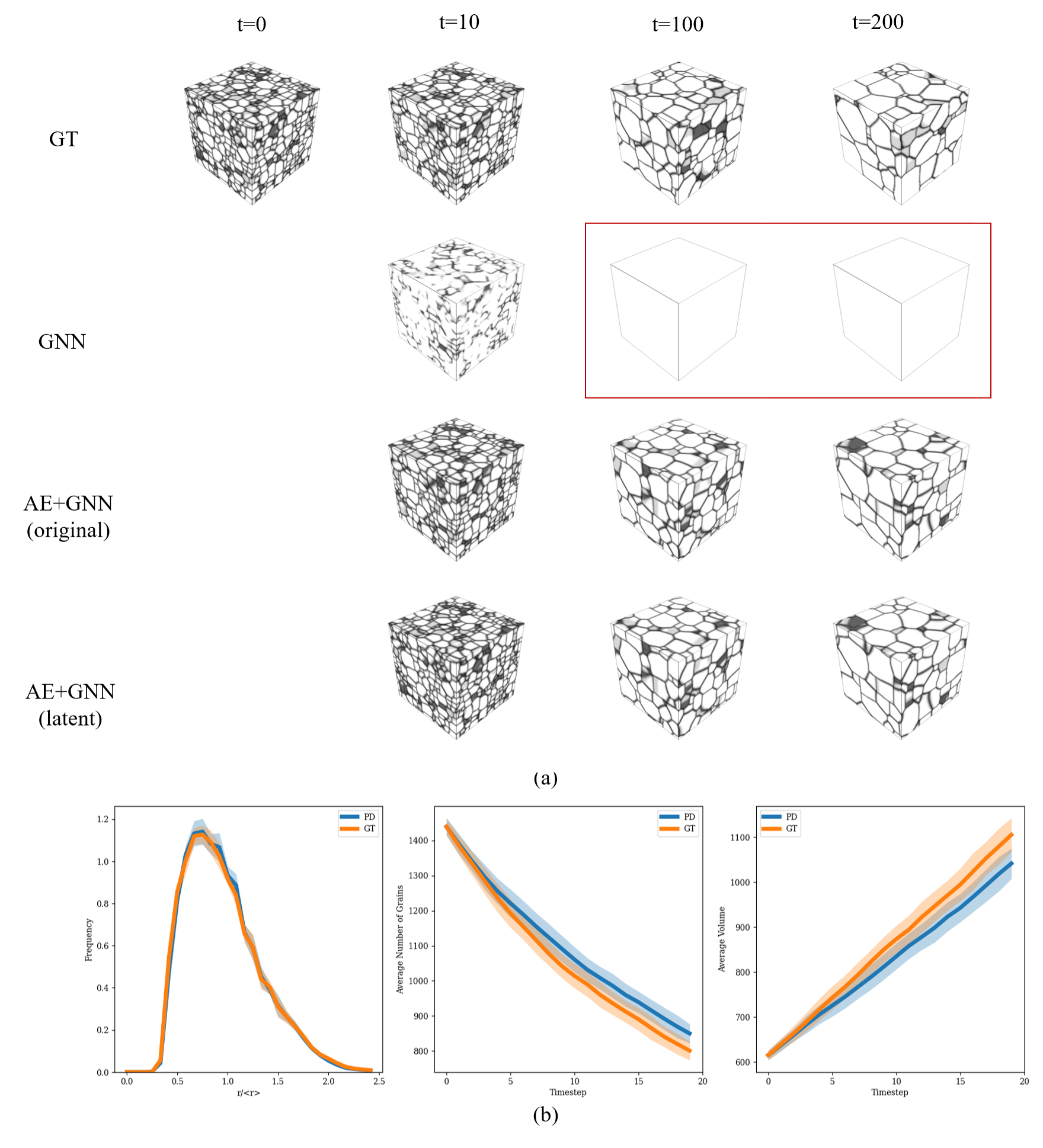}
  \caption{Spatiotemporal extrapolation and microstructure visualization in 3D. (a) Predictions trained on $32^3$ mesh with 25 frames and inferred on a $96^3$ mesh with 200 frames in GNN, AE+GNN (compression ratio 8 or $n=2$) in original and latent spaces {using 3 MP layers}, demonstrating spatiotemporal extrapolation. The GNN baseline performs only a few inference steps before divergence, indicated by the red rectangle. (b) Statistics of 6 independent predicted and
ground-truth trajectories on $96^3$ mesh for AE+GNN trained on $32^3$ mesh}
\label{fig:3dvisual}
\end{figure}

For 3D, the ground truth dataset consists of PMC trajectories of $N_t=25$  time frames of $32^3$ voxels, split into 726 training trajectories and 186 for validation. The model used 3 message passing layers and 5-step training strategy  of Eq.~(\ref{eq:multi-step-loss}).
%{predicting 5 output frames from 1 input frame.} 
We train the 3D model ($32^3$ mesh) with 25 frames and test its extrapolation ability for both spatial and temporal in ($96^3$ mesh) with 200 frames. A representative trajectory is shown in Figure~\ref{fig:3dvisual}(a), comparing the extrapolated predictions with the PMC ground truth and baseline GNN. The baseline GNN model can perform only a few inference steps before collapsing, whereas the AE+GNN model maintains consistency with the ground truth trajectories over long temporal horizons even when trained on a small mesh dataset. This demonstrates that the hybrid architecture effectively enhances the model’s spatiotemporal extrapolation capability. The statistics for AE+GNN extrapolation on $93^3$ is also tested correspondingly and shown in Fig.~\ref{fig:3dvisual}(b).

In general, the bijective autoencoder ensures the computational efficiency, especially for large mesh cases. The backbone GNN captures the dynamics within the latent space. Using multi-step training strategy explicitly incorporates temporal information which helps the long term inference behavior.

\subsection{Refinement of model architecture}

\begin{figure}[htbp]
  \centering
\includegraphics[width=\linewidth]{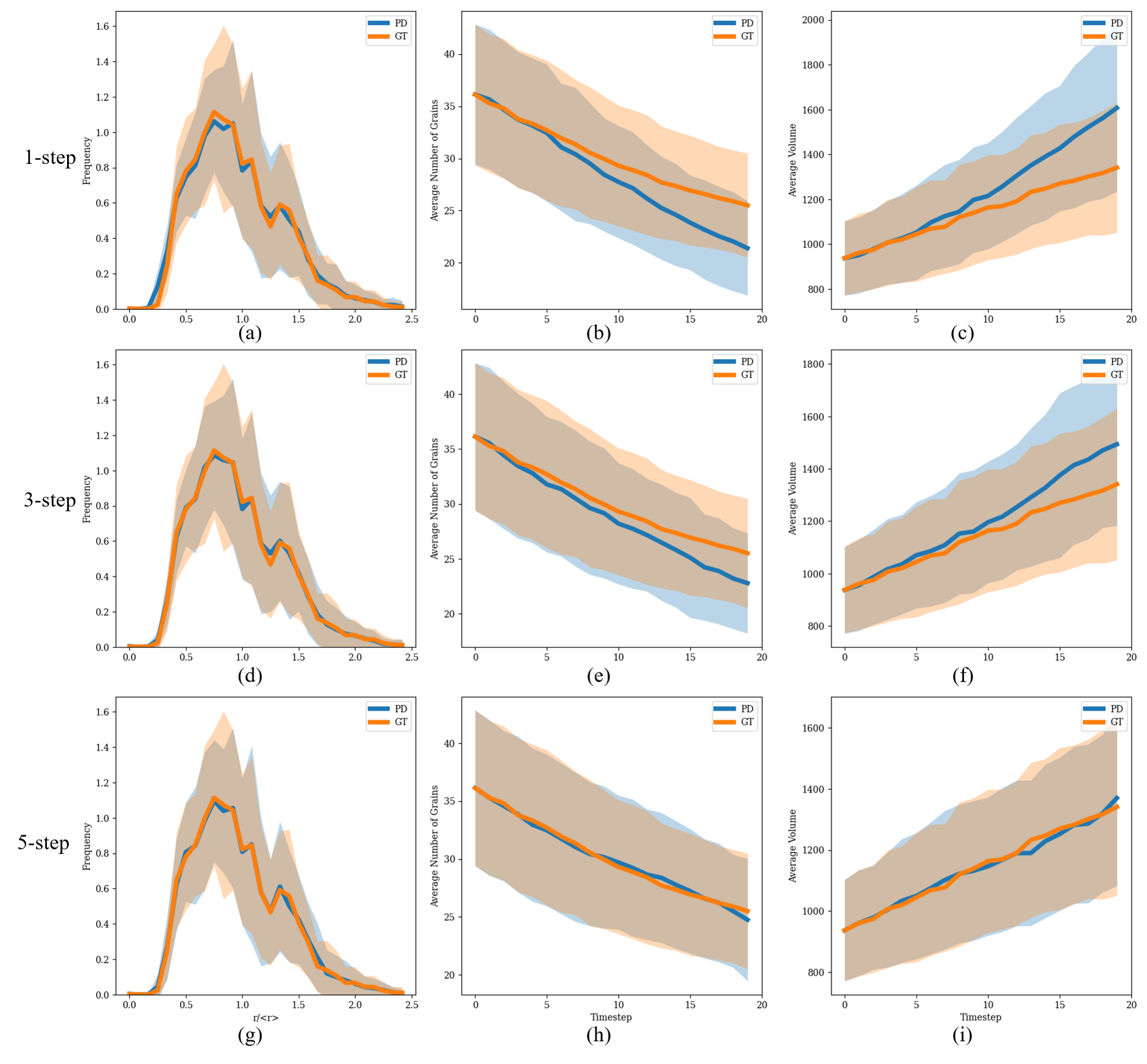}
  \caption{{Effects of multi-step training.} Statistical metrics of 3D grain simulations based on 40 independent predicted and ground-truth trajectories using a $32^3$ mesh. The first, second, and third rows correspond to models trained with next 1-step, 3-step, and 5-step, respectively. Columns represent the same metrics as in the Fig.~\ref{fig:nmp}}
\label{fig:statistic}
\end{figure}

\begin{figure}[htbp]
  \centering
\includegraphics[width=0.8\linewidth]{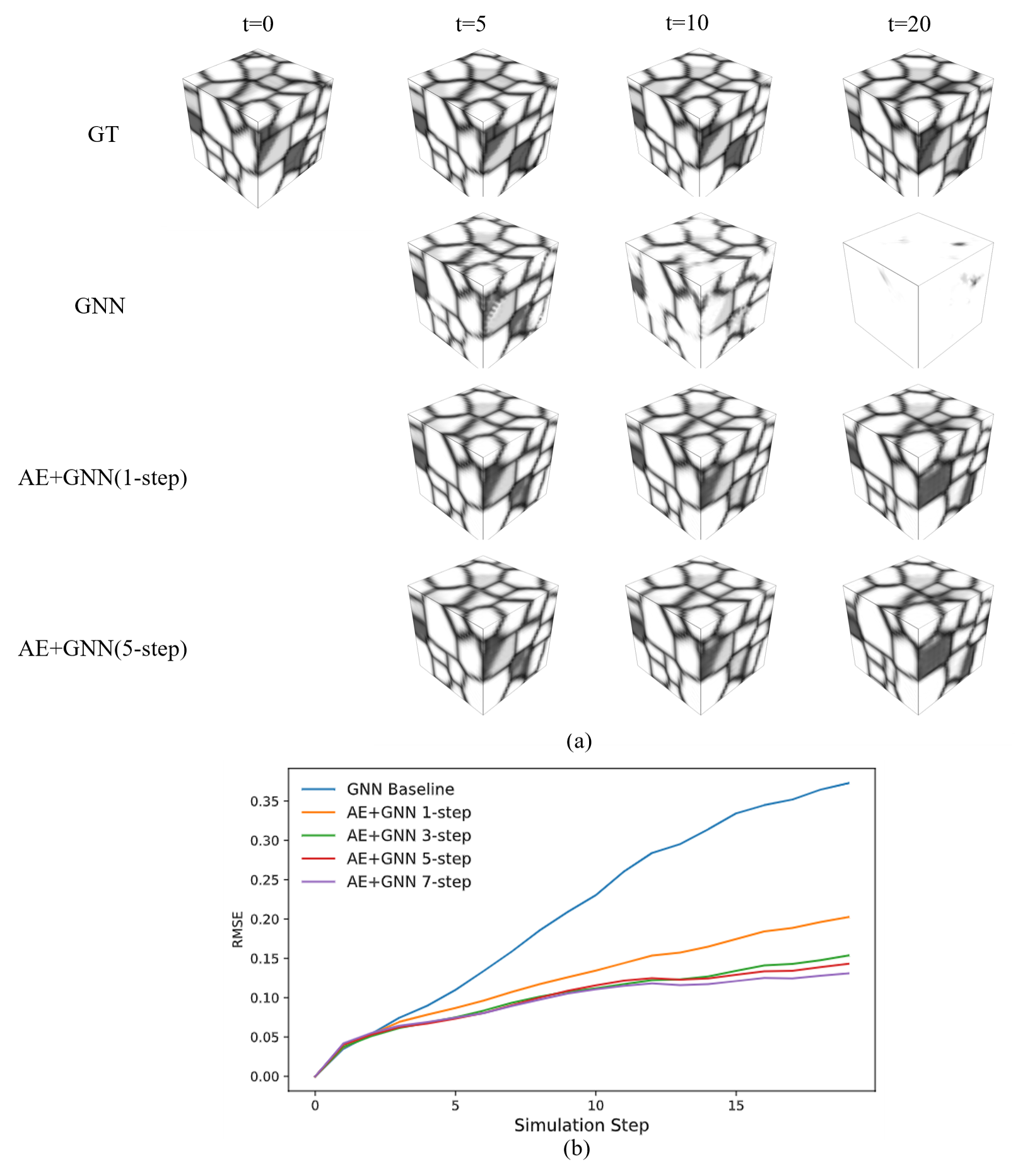}
  \caption{{Multi-step training result visualization and validation on $32^3$ mesh} (a) Microstructure visualization and (b) Inference RMSE versus simulation steps for GNN baseline and AE+GNN models, showing stabilization due to the multi-step loss.
  % }{RMSE for ground truth, inference based on GNN only trained model and AE+GNN with multi-step trained model.} {All models have 3 MP layers.}
  }
\label{fig:steps_acc}
\end{figure}

In addition to the key strategies discussed above, other hyperparameters were also investigated to evaluate their secondary effects on model accuracy and efficiency. 
We evaluate the importance of employing a multi-step strategy of Eq.~\ref{eq:multi-step-loss} during training. The result in Fig. \ref{fig:statistic}(a-c) shows the model using the next 1 step supervision during training produces reasonable short-term behavior but fails to capture the evolution dynamic of training data for long-term like grain growth speed shown in \ref{fig:statistic}(c). The multi-step training strategy effectively addresses this limitations by enforcing temporal coherence during training, leading to more stable long-term predictions. Fig.~\ref{fig:statistic}(d–f) and Fig.~\ref{fig:statistic}(g–i) show the results of models trained with the next 3-step and 5-step future steps, respectively. In these cases, the averages and envelopes of the statistical metrics from the ground truth and model inference become closer when more layers are used. The corresponding microstructure visualization and loss calculation are shown in Fig.~\ref{fig:steps_acc}(a) and (b). Note that our previous work trained on deterministic phase field data was able to capture both the short-term and long-term dynamics with 1 step supervision \cite{Fan2024MLST}. The need for multi-step supervision shown here is likely related to the noisy nature of the PMC data. More specifically, the $L_2$ loss applied to stochastic data will force a single-step training method to predict averaged blurry grain structures, while multi-step losses helps maintain clear boundaries.

\begin{figure}[htbp]
  \centering
\includegraphics[width=0.8\linewidth]{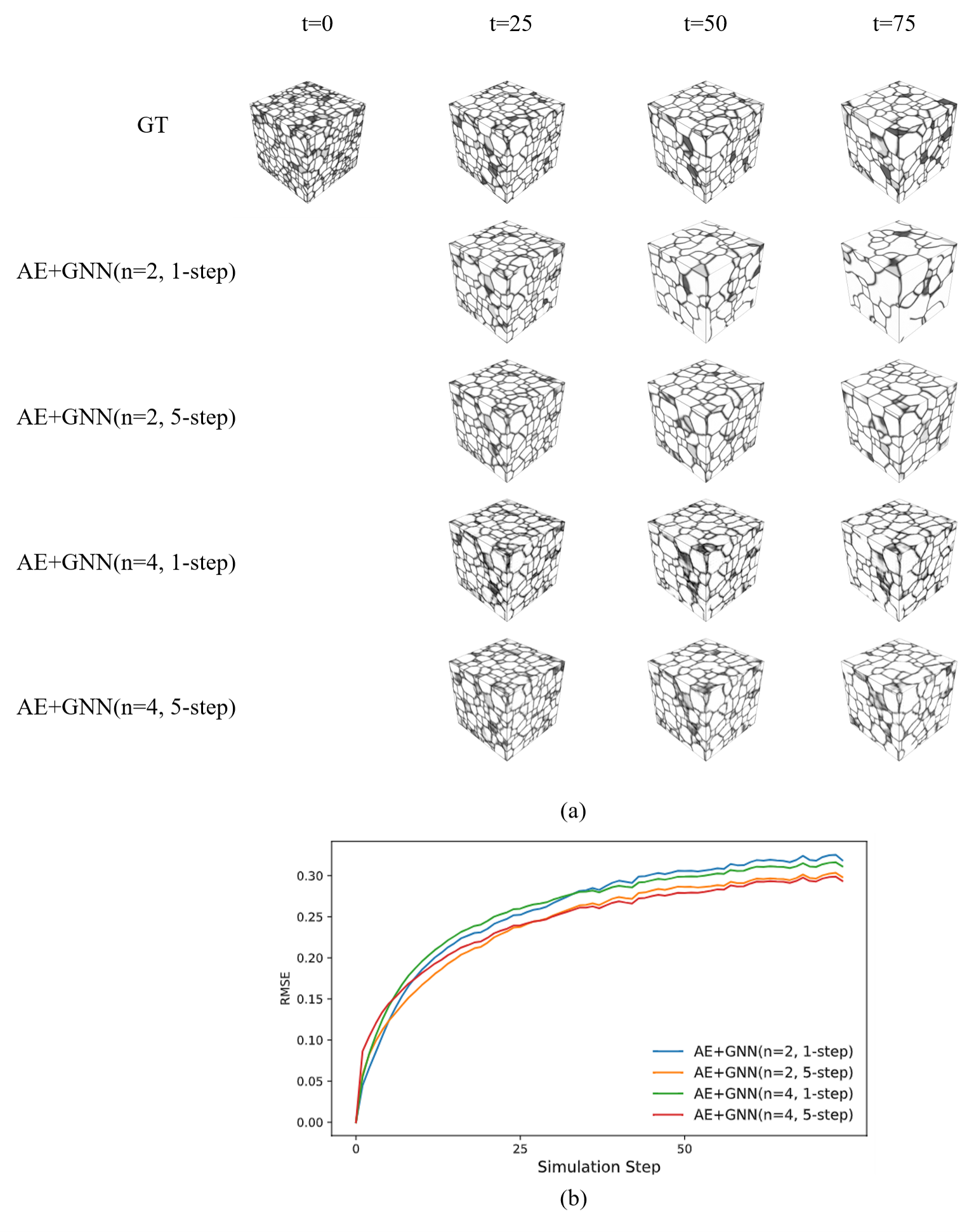}
  \caption{{Multi-step training result visualization and validation on $96^3$ with compression ratio 2(spatial ×8) and 4(spatial ×64).} (a) Microstructure visualization and (b) Inference RMSE versus simulation steps for GNN baseline and AE+GNN models, showing stabilization due to the multi-step loss.
  % }{RMSE for ground truth, inference based on GNN only trained model and AE+GNN with multi-step trained model.} {All models have 3 MP layers.}
  }
\label{fig:training_96}
\end{figure}

Similarly, we also test the multi-step strategy on the $96^3$ training case. Here, we rather than using the compression ratio 2 as training on $32^3$, we use compression ratio 2 and a higher compression ratio 4 for better computer efficiency.  Increasing the compression ratio from 2 to 4 offers substantially higher computational efficiency, as summarized in Table~\ref{tab:3d_inference} and even produces a slight improvement in performance shown in Fig.~\ref{fig:training_96}(b). In addition, we observe that the multi-step strategy boosts performance in all cases, independent of the chosen compression ratio. These results demonstrate that the improvement in computational efficiency provided by our method becomes much more significant for large meshes, while it can still maintain strong performance.

\begin{figure}[htbp]
  \centering
\includegraphics[width=\linewidth]{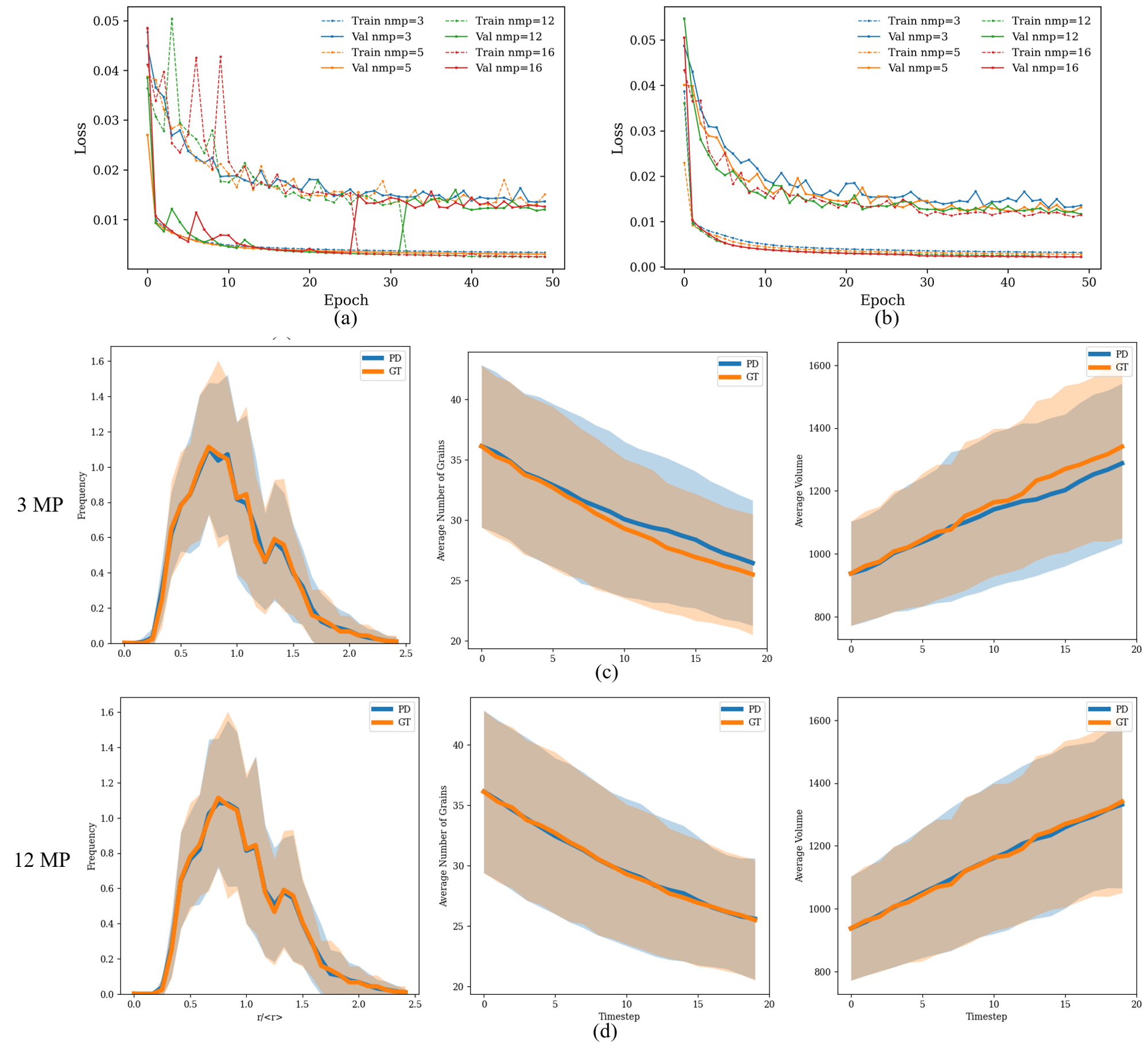}
  \caption{Effects of activation function and number of message-passing layers on  training of the AE+GNN model. (a) ReLU and (b) SiLU activation functions for 50 epochs. Statistical metrics of the inference trajectories using the {more stable} SiLU-trained model are shown in (c) for 3 MP layers and in (d) for 12 MP layers.}
\label{fig:relu_silu}
\end{figure}

Inspired by Ref.~\cite{wu2023demystifying}, we adopt the SiLU activation function~\cite{elfwing2018sigmoid} to replace ReLU in our GNN architectures to mitigate oversmoothing in AE+GNN model with large number of message passing layer shown in Fig.~\ref{fig:nmp}. The detailed training comparison is shown in Figure~\ref{fig:relu_silu}(a), where using SiLU helps maintain stability and mitigates performance degradation when the number of message-passing layers increases. The corresponding statistical plots  for inference are shown in Figure~\ref{fig:relu_silu}(d) where the ground truth and inference exhibit close agreement even using 12 message passing layer rather than performance degradation shown in Figure \ref{fig:nmp}(g-i).

Based on these designs, our model:

1. Significantly reduces memory usage, computational cost and number of passing layer required by GNN. These reductions becoming more substantial at larger spatial sizes, indicating strong computational scalability.

2. Accurately captures the stochastic characteristics of the PMC dataset.

3. Exhibits enhanced spatiotemporal modeling capability.

\section{Summary and conclusion}

Realistic grain boundary networks exhibit complex, evolving morphologies that cannot be captured by small simulation cells. Accurate modeling therefore requires training and inference on large-scale grain structures, and hence scalable architectures capable of handling large microstructural data efficiently. It is therefore important to reduce the memory usage and increase the spatial scales of the simulations, and to reduce runtime. In this work, we develop mesoscale surrogate models with a novel hybrid CNN-GNN framework for grain growth trained from stochastic Potts Monte Carlo simulations. By incorporating dimensionality reduction through a bijective autoencoder, the model can be trained efficiently on large spatial datasets to learn microstructural correlations at more realistic scales. Substantial reduction in inference memory usage and moderate speedup were achieved. The reduction in computational cost becomes more pronounced as the spatial size increases, indicating that the proposed method is highly scalable. The largest case in our study ($160^3$) achieves over a hundredfold reduction in both memory usage and runtime compared to the GNN-only baseline. The latent space inference strategy leads to further runtime reduction.
Currently the best performance is achieved with moderate linear compression ratios (e.g., 2–4). Another important advantage is that dimension reduction decreases the number of message-passing layers in the GNN required for accurate learning of the coarsening dynamics from 12 in the baseline GNN model to 3 in the new hybrid one.
{The CNN-based bijective autoencoder not only extracts effective features for the GNN but also enhances prediction accuracy.} In contrast to previous surrogate models trained with a loss function against single next step in a deterministic dataset, a multi-step loss was adopted to learn from a stochastic PMC dataset. Overall, our approach highlights a scalable strategy for applying neural networks to study grain growth at large spatial scales with improved computational efficiency and accuracy.

Future work will focus on applying this approach to real experimental datasets. Although our approach performs robustly on stochastic simulation datasets, real experimental data are typically more complex and much more limited in quantity than the datasets used for training. Integrating previously developed adaptive mesh refinement techniques for graph neural networks \cite{Fan2024MLST} may help address these challenges. 
Although the training data are generated from stochastic KMC simulations, 
the present CNN--GNN model produces deterministic predictions during inference, 
effectively predicting the mean time evolution. 
Incorporating stochasticity into the learned evolution model is the subject of future work, which requires careful modeling of not just the predicted mean, but covariance matrix that must be consistent with the simulation temperature. This line of interesting work will likely involve more physics constraints rather than a pure data-driven approach.

\section*{Acknowledgement}
ES was supported by the Critical Materials Innovation Hub, an Energy Innovation Hub funded by the U.S. Department of Energy, Office of Energy Efficiency and Renewable Energy, and Advanced Materials and Manufacturing Technologies Office. We acknowledge support by the Laboratory Directed Research and Development (LDRD) program (25-ERD-002 for FZ, 22-ERD-016 for VB) at Lawrence Livermore National Laboratory (LLNL). 
This work was performed under the auspices of the U.S. Department of Energy by LLNL under contract DE-AC52-07NA27344.

\section*{Author Contributions}
F.Z. conceived and supervised the project and implemented the model. E.S. performed PMC simulations. E.S. and Z.T. carried out the computational experiments and analyzed the data. E.S., Z.T. and F.Z. wrote the manuscript with inputs from other authors.

\section*{Code Availability}
The code will be made publicly available upon publication.

\section*{Declaration of Interests}
The authors declare no competing interests.

\bibliographystyle{elsarticle-num}
\bibliography{refs}

\clearpage
\appendix
% \section{Figures}

% Supporting Information Appendix
\newpage
\appendix
\counterwithin{figure}{section}
\counterwithin{table}{section}
\counterwithin{equation}{section}

% Reset all counters for Supporting Information
\setcounter{figure}{0}
\setcounter{table}{0}
\setcounter{equation}{0}
\setcounter{page}{1}

% Redefine figure, table, and equation labels for Supporting Information
\renewcommand{\thefigure}{S\arabic{figure}}
\renewcommand{\thetable}{S\arabic{table}}
\renewcommand{\theequation}{S\arabic{equation}}
\renewcommand{\thepage}{S\arabic{page}}

% Supporting Information title
\section*{Supporting Information}
% \addcontentsline{toc}{section}{Supporting Information}

% Optional: Add a note about the Supporting Information
\noindent \textbf{Note:} This Supporting Information contains additional data, figures, and detailed procedures that supplement the main text.

\subsection*{Additional Details}

\begin{figure}[htbp]
    \centering
    \includegraphics[width=\columnwidth]{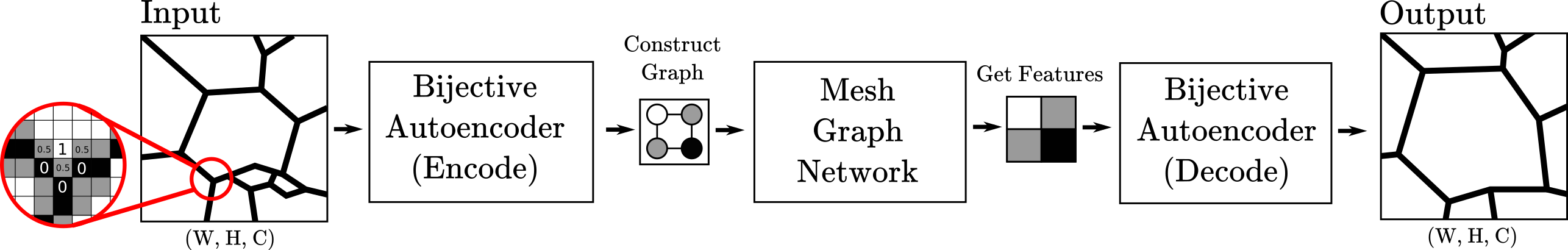}
    \includegraphics[width=\columnwidth]{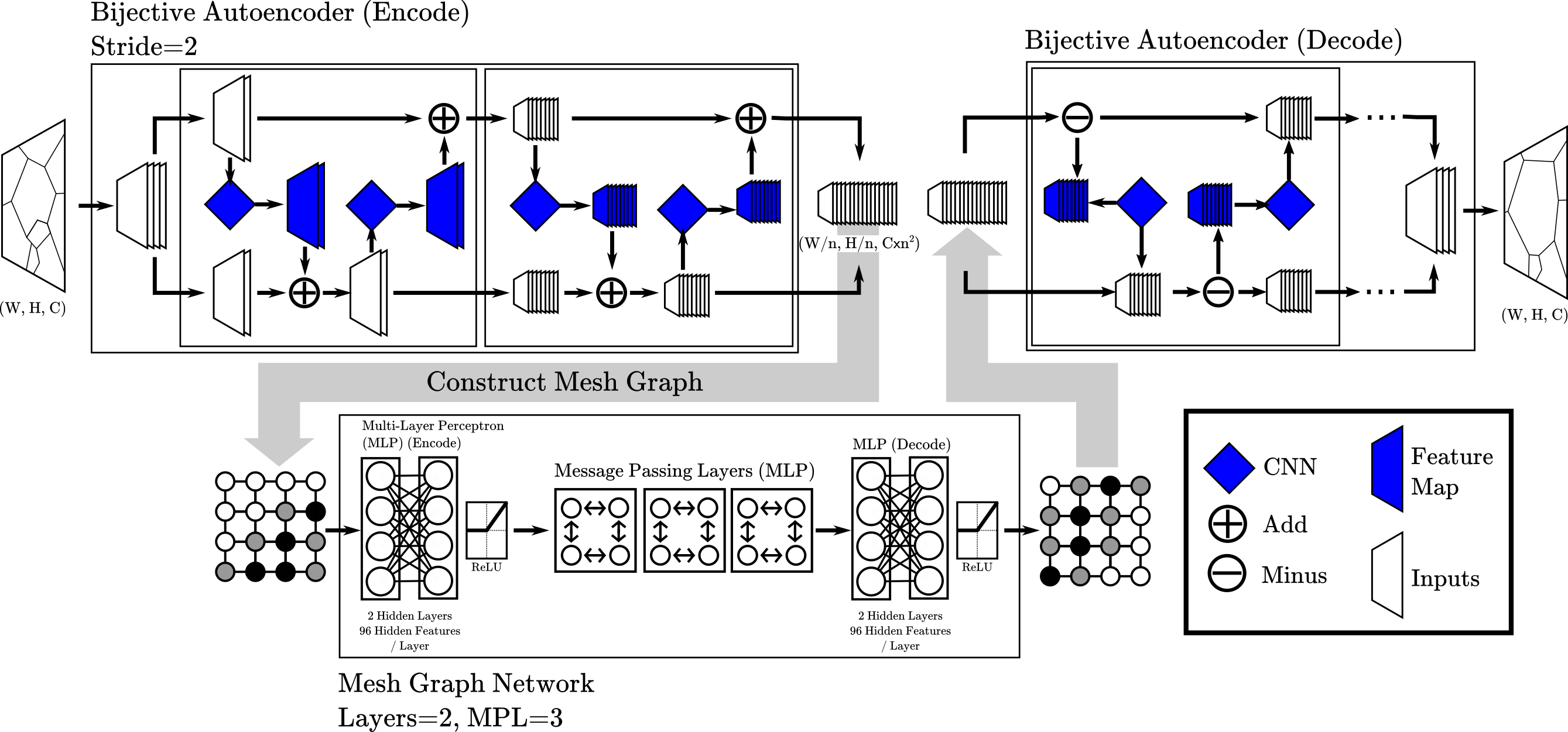}
    \caption{(Top) Overview of the neural network surrogate model. (Bottom) Detailed breakdown of the surrogate model 2-way autoencoder and Mesh GNN for a configuration of Stride=2, Layers=2, MPL=3. }
    \label{fig:NN-concept}
\end{figure}

\begin{figure}[htbp]
    \centering
    \includegraphics[width=\columnwidth]{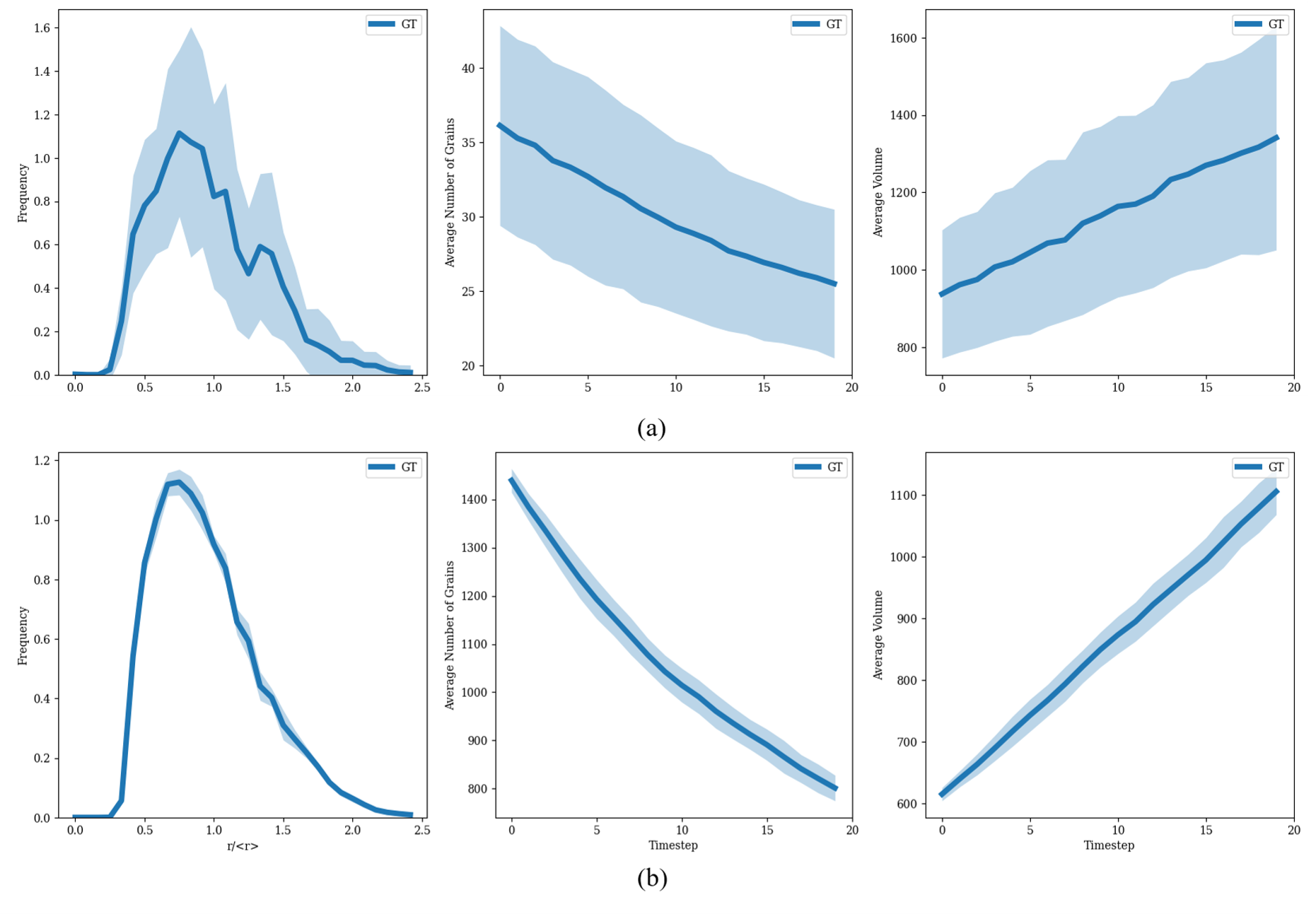}
    \caption{Statistical of 3D training datasets: (a) 40 independent $32^3$. (b) 6 independent $96^3$. From left to right, the columns show the normalized grain diameter distribution, the number of grains, and the average grain area as a function of time. The solid line represents the mean value, and the shaded region denotes the deviation across independent simulations.}
    \label{fig:shared_grains}
\end{figure}

\begin{table}[t]
\centering
\small
\setlength{\tabcolsep}{5pt}
\renewcommand{\arraystretch}{1.2}

\makebox[\textwidth][c]{%
\begin{threeparttable}
\begin{tabular}{l c c c c c c c}
\toprule
\textbf{Memory} & 
\textbf{GNN only} & \multicolumn{3}{c}{\textbf{AE+GNN}} \\ 
\midrule
$320^2$ & 6.45
 & 1.64
 & 0.42
 & 0.12
 \\

$640^2$ & 25.79
 & 6.50
 & 1.69
 & 0.48
 \\
 
$1280^2$ & \cellcolor{red!30}103.13
 & 25.97
 & 6.69
 & \cellcolor{blue!20}1.90
 \\

\midrule
\textbf{Runtime} & 
\textbf{GNN only} & \multicolumn{3}{c}{\textbf{AE+GNN}}\\

\midrule
$320^2$ & 9.36e+01
 & 5.00e+01
 & 6.68e+01
 & 9.21e+01
 \\

$640^2$ & 3.24e+02
 & 1.07e+02
 & 7.03e+01
 & 8.55e+01
 \\
 
$1280^2$ & \cellcolor{red!30}1.22e+03
 & 3.72e+02
 & 1.45e+02
 & \cellcolor{blue!20}1.04e+02

 \\

\bottomrule
\end{tabular}
\end{threeparttable}
}% end of makebox
\caption{Memory comparison(GB) and runtime(s) for large meshes in 2D training. Columns under AE+GNN reprsent linear compression ratio 2, 4, 8 from left to right. Highlighted cells mark the largest mesh case, comparing the GNN-only baseline against the AE+GNN model with a compression ratio of 8.}
\label{tab:2d_train}
\end{table}

\begin{table}[t]
\centering
\small
\setlength{\tabcolsep}{5pt}
\renewcommand{\arraystretch}{1.2}

\makebox[\textwidth][c]{%
\begin{threeparttable}
\begin{tabular}{l c c c c c c c}
\toprule
\textbf{Memory} & 
\textbf{GNN only} & \multicolumn{3}{c}{\textbf{AE+GNN}} \\ 
\midrule
$64^3$ & 69.040
 & 8.863
 & 1.342
 & 0.515
 \\

$96^3$ & \cellcolor{red!30}OOM
 & 29.896
 & 4.521
 & \cellcolor{blue!20}1.649

 \\
 
$128^3$ & \cellcolor{red!30}OOM
 & 70.838
 & 10.670
 & \cellcolor{blue!20}3.786
 \\

\midrule
\textbf{Runtime} & 
\textbf{GNN only} & \multicolumn{3}{c}{\textbf{AE+GNN}}\\

\midrule
$64^3$ & 1.12e+03
 & 3.49e+02
 & 2.34e+02
 & 2.62e+02

 \\

$96^3$& \cellcolor{red!30}---
 & 1.03e+03
 & 7.59e+02
 & \cellcolor{blue!20}7.55e+02

 \\
 
$128^3$ & \cellcolor{red!30}---
 & 2.48e+03
 & 1.90e+03
 & \cellcolor{blue!20}1.75e+03

 \\

\bottomrule
\end{tabular}
\end{threeparttable}
}% end of makebox
\caption{Memory comparison(GB) and runtime(s)  for large meshes in 3D training. Columns under AE+GNN reprsent linear compression ratio 2, 4, 8 from left to right. Red cells indicate GNN-only runs that exceed memory limits (out-of-memory), while blue cells highlight AE+GNN with a compression ratio of 8 successfully complete the simulations with a very low memory usage.(–- indicates OOM failure)}
\label{tab:3d_train}
\end{table}

\begin{table}[t]
\centering
\small
\setlength{\tabcolsep}{5pt}
\renewcommand{\arraystretch}{1.2}

\makebox[\textwidth][c]{%
\begin{threeparttable}
\begin{tabular}{l c c c c c c c}
\toprule
\textbf{Memory} & 
\textbf{GNN only} & \multicolumn{3}{c}{\textbf{AE+GNN(original)}}  & \multicolumn{3}{c} {\textbf{AE+GNN(latent)}} \\ 
\midrule
$1024^2$ & 15.78
 & 4.23
 & 1.33
 & 0.60
& 3.98
& 1.08
&0.35

 \\

$2048^2$ & 63.12
 & 16.90
 & 5.30
 & 2.39
& 15.90
& 4.29
&1.39

 \\
 
$2688^2$ & \cellcolor{red!30}108.73
 & 29.13
 & 9.13
 & 4.14
& 27.39
& 7.39

&\cellcolor{blue!30}2.40

 \\

\midrule
\textbf{Runtime} & 
\textbf{GNN only} & \multicolumn{3}{c}{\textbf{AE+GNN(original)}}  & \multicolumn{3}{c} {\textbf{AE+GNN(latent)}}\\

\midrule
$1024^2$ & 1.591e+01
 & 8.149e+00
 & 2.004e+00
 & 1.720e+00
& 8.256e+00
& 3.015e+00
&7.322e-01

 \\

$2048^2$ & 2.799e+01
 & 1.570e+01
 & 7.145e+00
 & 3.717e+00
& 1.616e+01
& 4.037e+00
&1.283e+00

 \\
 
$2688^2$ & \cellcolor{red!30}7.626e+01
 & 1.913e+01
 & 7.902e+00
 & 6.457e+00
& 1.838e+01
& 5.624e+00

& \cellcolor{blue!30} 1.790e+00

 \\

\bottomrule
\end{tabular}
\end{threeparttable}
}% end of makebox
\caption{Memory comparison(GB) and runtime(s) and  for large meshes in 2D inference. Columns under AE+GNN(orignal) and AE+GNN(latent) reprsent linear compression ratio 2, 4, 8 from left to right. Highlighted cells mark the largest mesh case, comparing the GNN-only baseline against the AE+GNN(latent) model with a compression ratio of 8.}
\label{tab:2d_inference}
\end{table}

\begin{figure}[htbp]
  \centering
\includegraphics[width=\linewidth]{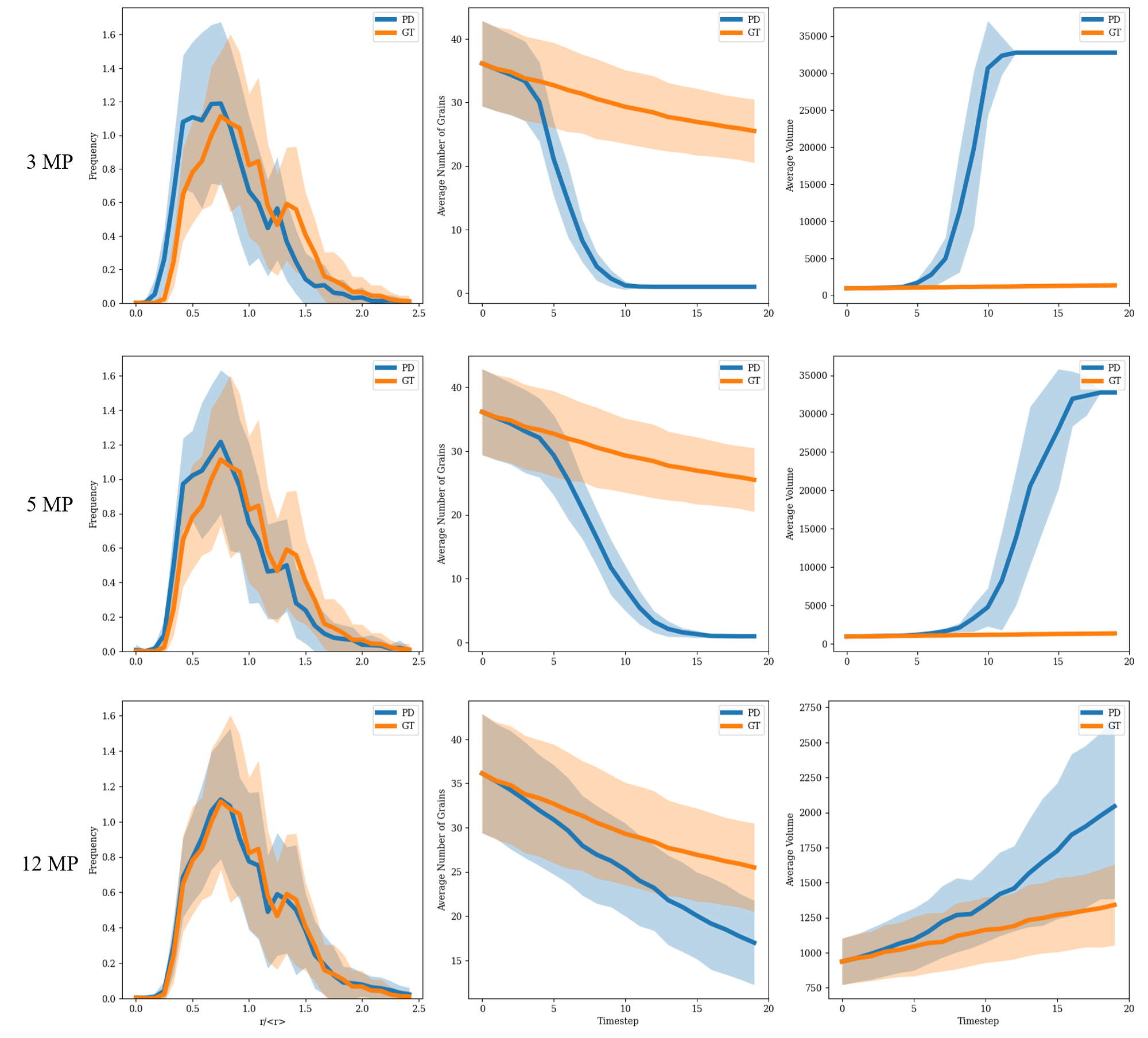}
  \caption{{Effects of number of message passing layer on GNN only model.} Statistical metrics of 3D grain simulations for GNN only based on 40 independent predicted and ground-truth trajectories using a $32^3$ mesh. The first, second, and third rows correspond to models using 3, 5, 12 message passing layers in GNN, respectively. Columns represent the same metrics as in the Fig.\ref{fig:nmp}.}
\label{fig:gnn_nmp}
\end{figure}

% \begin{table}[t]
% \centering
% \small
% \begin{threeparttable}
% \begin{tabular}{lcccc}
% \toprule
% \textbf{Runtime} & \textbf{KMC} & \textbf{GNN only} & \textbf{AE+GNN (original)} & \textbf{AE+GNN (latent)} \\
% \midrule
% $512^2$ & \cellcolor{red!20}1.453e+03 & 1.248e+00 & 2.750e-01 & \cellcolor{blue!15}8.318e-02 \\
% $96^3$  & \cellcolor{red!20}1.629e+02 & 6.554e+00 & 4.687e-01 & \cellcolor{blue!15}1.546e-01 \\
% $128^3$ & \cellcolor{red!20}4.037e+02 & 1.270e+01 & 8.116e-01 & \cellcolor{blue!15}7.581e-01 \\
% $160^3$ & \cellcolor{red!20}8.099e+02 & 1.269e+02 & 2.236e+00 & \cellcolor{blue!15}1.104e+00 \\
% \bottomrule
% \end{tabular}
% \end{threeparttable}
% \caption{Runtime (s) for large meshes in 2D and 3D inference. KMC simulations are executed on a CPU node using 36 MPI ranks, whereas neural-network inference is performed on a single AMD MI300A GPU. Columns under AE+GNN (original) and AE+GNN (latent) represent models using a linear compression factor of 8. Highlighted cells mark the largest mesh case comparing KMC and AE+GNN (latent).}
% \label{tab:KMC_compare}
% \end{table}

\begin{table}[t]
\centering
\small
\begin{threeparttable}
\begin{tabular}{lcccc}
\toprule
\textbf{Runtime} & \textbf{KMC} & \textbf{GNN only} & \textbf{AE+GNN (original)} & \textbf{AE+GNN (latent)} \\
\midrule

$96^3$  & \cellcolor{red!20}1.629e+02 & 6.554e+00 & 4.687e-01 & \cellcolor{blue!15}1.546e-01 \\
$128^3$ & \cellcolor{red!20}4.037e+02 & 1.270e+01 & 8.116e-01 & \cellcolor{blue!15}7.581e-01 \\
$160^3$ & \cellcolor{red!20}8.099e+02 & 1.269e+02 & 2.236e+00 & \cellcolor{blue!15}1.104e+00 \\
\bottomrule
\end{tabular}
\end{threeparttable}
\caption{Runtime (s) for large meshes in 3D inference. KMC simulations are executed on an Intel Xeon Platinum 8480+ CPU node using 36 MPI ranks, whereas neural-network inference is performed on a single AMD MI300A GPU. Columns under AE+GNN (original) and AE+GNN (latent) represent models using a linear compression factor of 8. Highlighted cells mark the largest mesh case comparing KMC and AE+GNN (latent).}
\label{tab:KMC_compare}
\end{table}

\end{document}